\documentclass[]{article}
\usepackage[a4paper, total={7in, 10in}]{geometry}
\usepackage[utf8]{inputenc}
\usepackage[T1]{fontenc}
\usepackage[]{biblatex}
\usepackage{amssymb}
\usepackage{graphicx}
\usepackage{hyperref}
\usepackage{float}
\usepackage{subcaption}

\newif\ifsetDraft
\setDrafttrue
\ifsetDraft
	\usepackage[colorinlistoftodos]{todonotes}
	\newcommand{\jakub}[1]{\todo[author=jakub,size=\small,inline,color=green!40]{#1}}
    
\else
	\newcommand{\jakub}[1]{}
	
\fi

\author{ \large
M. Ja\'skowski\textsuperscript{$\ast\dagger$} \hspace{0.5cm} J. \'{S}wi\k{a}tkowski\textsuperscript{$\ast\dagger$} \hspace{0.5cm} M. Zaj\k{a}c\textsuperscript{$\ast\dagger$}\\
\large M. Klimek \textsuperscript{$\dagger$} \hspace{0.5cm} J. Potiuk \textsuperscript{$\dagger$} \hspace{0.5cm} P. Rybicki \textsuperscript{$\dagger$} \hspace{0.5cm} P. Polatowski \textsuperscript{$\dagger$} \\
\large P. Walczyk \textsuperscript{$\dagger$} \hspace{0.5cm} K. Nowicki \textsuperscript{$\dagger$} \hspace{0.5cm} M. Cygan \textsuperscript{$\dagger\ddagger$} \\\\
\large	\textsuperscript{$\dagger$}NoMagic.AI\\
\large	\textsuperscript{$\ddagger$}Institute of Informatics, University of Warsaw\\ \\
\normalsize \textsuperscript{$\ast$} MJ, JS, and MZ contributed equally to this work
}
\title{\textbf{Improved GQ-CNN: Deep Learning Model for Planning Robust Grasps}}

\begin{document}
\maketitle


\section{Abstract}
Recent developments in the field of robot grasping have shown
great improvements in the grasp success rates when dealing with unknown objects.
In this work we improve on one of the most promising approaches, the Grasp Quality Convolutional Neural Network (GQ-CNN) trained on the DexNet 2.0 dataset \cite{mahler2017dex}.
We propose a new architecture for the GQ-CNN 
and describe practical improvements
that increase the model validation accuracy from 92.2\% to 95.8\%
and from 85.9\% to 88.0\% on respectively image-wise and object-wise training and validation splits.
\section{Background}

\paragraph{Robot grasping.} Robot grasping is one of the most important tasks on the way to a wide adoption of
robots at our homes and in the industry. However, grasping is also one of the most challenging tasks
to solve.

Robot grasping methods can be categorized based on the type of grasp success
criteria as either analytical \cite{prattichizzo2016grasping,pokorny2013classical,rodriguez2012caging} or empirical (data-driven) \cite{bohg2014data}.
In analytical methods, physics simulations are used to precompute
the grasp robustness scores based
on predefined 3D models and the most robust grasp for the best matching 3D model is executed \cite{brook2011collaborative,hernandez2016team,kehoe2013cloud,mahler2016dex}.
 In empirical methods, the grasp
success is predicted directly from the robot sensors, typically using machine learning models
\cite{bohg2014data,levine2016learning}.
The empirical methods are usually a lot faster and generalize better to variations in object shapes
and to unseen objects than the analytical methods.
However, the empirical methods often require a large number of labels to train.
Such labels can be either obtained
from tedious human labeling \cite{detry2013learning,kappler2015leveraging,lenz2015deep,herzog2014learning,redmon2015real} or 
long running physical trials \cite{pinto2016supersizing,pinto2017supervision,levine2016learning}.



\paragraph{Grasp Quality CNN}
The work by \textcite{mahler2017dex} that forms the basis for our work is an empirical method
for robot grasping. They train a CNN on a large dataset of physics-based generated labels to
predict the grasp success. More precisely, they generate a Dex-Net 2.0 dataset that contains 6.7
million point clouds and analytical grasp success metrics for parallel-jaw grasps
based on a dataset of 1,500 3D object models.
They devise a Grasp Quality CNN (GQ-CNN) network for predicting the grasp success
based on the 2.5D point clouds from a depth camera. Finally, they develop a grasp planning policy
that samples grasps and ranks them using the trained GQ-CNN. Their solution provides very promising
results that is comparable in grasp success with the state-of-the-art analytical methods, but
outperforms them for unseen objects and is $3\times$ faster. 

\section{GQ-CNN improvements}

Encouraged by the success of the original work by \textcite{mahler2017dex} we propose
improvements to their GQ-CNN network.
Our improvements come in two forms. First, we propose a new CNN architecture that
achieves higher test accuracy on synthetic data. Second, we improve their data augmentation
procedure, which results is an increased robustness of the model. 

\subsection{Problem statement}

We are interested in planning a robust planar parallel-jaw grasp of a single object on a table
from a 2.5D point cloud. GQ-CNN learns a function that takes as input $32\times32$ depth images
and a grasp depth $z \in \mathcal{R}$ and returns the grasp success probability.
The planar position of the grasp is defined by the center of the input image
and the orientation of the gripper $\varphi$ is parallel to the horizontal axis of the image.

\subsection{New GQ-CNN architecture}
\label{magic_GQ-CNN}

\begin{figure}[H]
\centering
\includegraphics[width=0.8\textwidth]{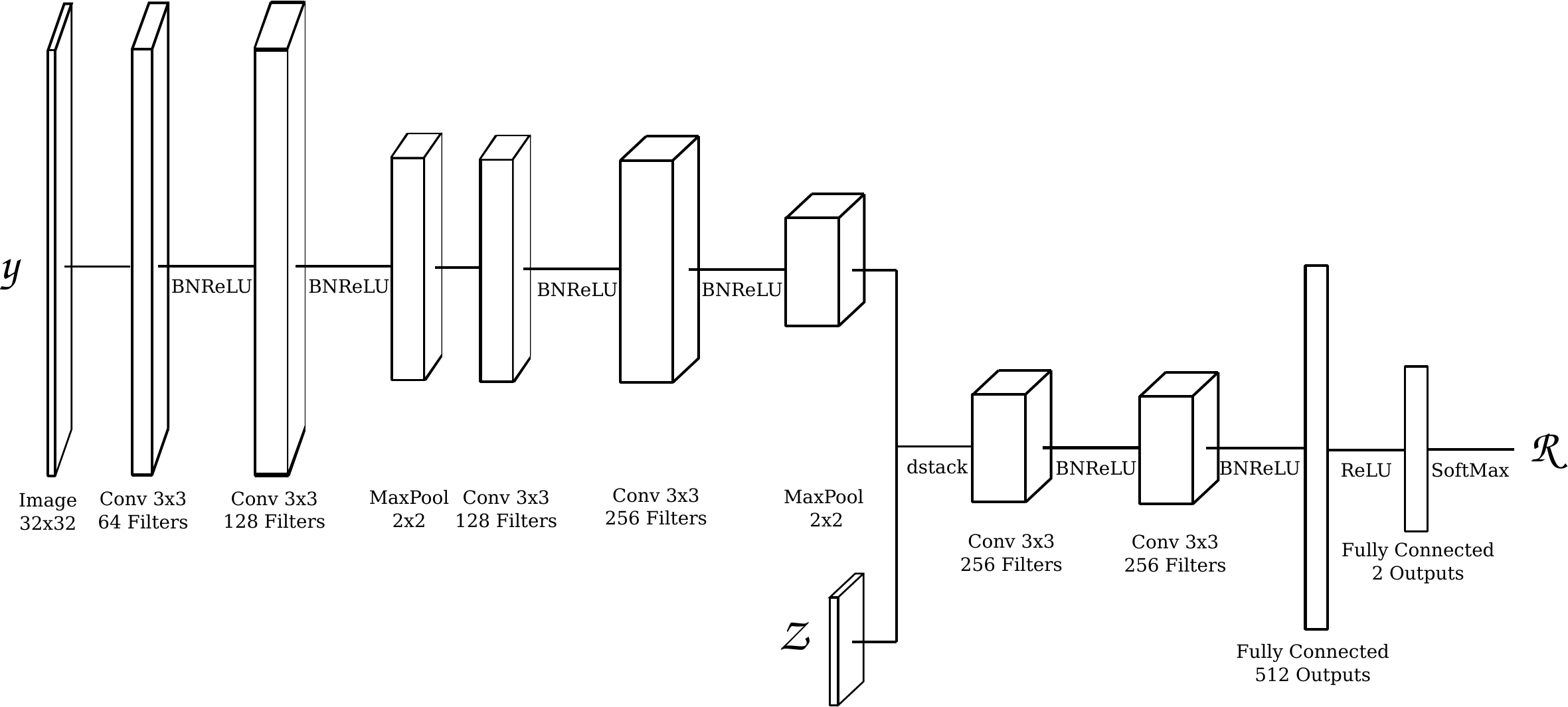}
\caption{Our GQ-CNN architecture.}
\label{fig:model_main_arch}
\end{figure}

The main contribution of this paper is the new architecture for the GQ-CNN (see \autoref{fig:model_main_arch}). 
We highlight the key discrepancies between our architecture and the original GQ-CNN architecture. First, instead of merging the
image and the grasp depth towers using a fully connected layer, we combine the two with a
convolutional layer.
To achieve that we reshape the grasp depth value $z$ to match the $2D$ shape of the
corresponding convolution layer in the tower for the image input.
The convolutional layer, as opposed to the fully connected layer, persists the spatial
 information even after the depth of the grasp is known.
 We also add two additional convolutional layers after the two towers are merged.
We hypothesize that this allows the model to focus its attention on the spatial locations with heights
similar to the grasp height. 
Furthermore, we increase the number of filters 
and add an additional max pooling operation after the last convolutional layer in the image tower.
This results in a deeper and more complex network architecture than the original one.
Finally, instead of using the Local Response Normalization (LRN)
 \cite{krizhevsky2012imagenet} after the second and fourth convolutional layer, we
 employ Batch Normalization \cite{ioffe2015batch} after each of the convolutional
layers. 

\subsection{Data augmentation}
\label{sec:data_aug}

Our second contribution is the analysis and improvement of the data augmentation procedure.
In particular, we investigate the impact of the data augmentation 
 on the test accuracy of
the model (on synthetic data). 
Below we describe the augmentations from the original work
(symmetrization, multiplicative aug. and Gaussian Process aug.) and propose a modification
to the multiplicative augmentation where we additionally adjust the grasp depth $z$.
Separately from the below augmentations, all the input images are normalized by subtracting their pixel mean and dividing by
their standard deviation ('Normalize').


\paragraph{Symmetrize} Images are flipped vertically or horizontally with 50\% probability.

\paragraph{Multiplicative augmentation} All pixels in an image are multiplied by a random variable
from a univariate Gamma distribution:
$$ f(x, a, b) = \frac{x^{a - 1}e^{-\frac{x}{b}}}{b^a\gamma(a)}$$
where shape $a = 1000$ and scale $ b = \frac{1}{a} = 0.001$. This results in multiplying the image by
a random noise highly concentrated around $\mu=1$ ($\sigma = 0.0316$). 

\paragraph{Gaussian Process augmentation} An additive augmentation applied with 50\% probability, where a Gaussian
Process (GP) noise is added to the pixel values. The GP noise is simulated by generating
a matrix of size $8\times8$ with values from a univariate Gaussian with $\mu = 0$ and $\sigma= 0.005$
and upsampled using bicubic interpolation. 

\paragraph{Multiplicative augmentation with grasp depth $z$ adjustment}
While the above multiplicative augmentation
effectively changes the height of the scene in the image, it
neither adjusts accordingly the grasp depth nor does it change the label grasp quality.
Therefore, we propose to adjust the grasping height by the exact same multiplicative augmentation as done to the pixels so that
we obtain a new example with a correct label as calculated for the image before the augmentation.

\section{Experiments}

\subsection{Experimental setup}


We perform our experiments on the same dataset training/validation splits as used in \cite{mahler2017dex}.
In particular, we report our results on three different splits of training and validation datasets:
\begin{itemize}
\item Object split: based on the unique objects (same object cannot be present in training and validation). 
\item Pose split: based on the unique poses of objects
(same object can be present in both training and validation)
\item Image split: based on the unique grasps (poses and objects can mix between the training and validation dataset).
\end{itemize}

In the experiments we use our GQ-CNN architecture as described in \autoref{magic_GQ-CNN}.
Optimization is performed using an Adam optimizer \cite{kingma2014adam} with learning
rate $10^{-4}$ 
and an exponential decay rate of $0.95$ every $50000$ steps,
a weight decay of $10^{-5}$ and a batch size of $128$.


\subsection{Improved GQ-CNN results}

\autoref{tab:splits_tab_nomagic} compares the results of our version of the GQ-CNN with the version
 from \textcite{mahler2017dex}. Our GQ-CNN network outperforms the baseline
in the image split and achieves comparable results on the pose and object splits.
We attribute the improvement on the image split to the higher expressiveness of our GQ-CNN (larger number of weights),
which results in an better fit to the training data (see 'Normalize' in \autoref{fig:image_split_plot}).
In the image split, the training data is very similar to the
 test data and thus our GQ-CNN
achieves much better performance also on the validation data
\footnote{The result for the image splits was also replicated by the authors of the original
GQ-CNN model \cite{mahler2017dex} and can be found on their leaderboard.
See: \url{https://berkeleyautomation.github.io/gqcnn/benchmarks/benchmarks.html}.}.
This does not hold for the other two
splits, where the training and validation sets are more differentiated and our GQ-CNN
reports no improvement over the baseline.

\begin{table}[H]
\centering
\begin{tabular}{lccc}
              & \multicolumn{3}{c}{Train/validation split type} \\ \cline{2-4}
              & Image          & Pose          & Object         \\ \hline
\textcite{mahler2017dex} & 92.2           & 88.9          & 85.9           \\
Ours (no data aug.) & 96.7           & 88.2          & 86.7
\end{tabular}
\caption{Validation accuracy of the GQ-CNN on different types of splits.
}
\label{tab:splits_tab_nomagic}
\end{table}

\subsection{Data augmentation results}

We investigate the impact of the data augmentation on the model performance and generalization,
but also on the calibration of its predictions.

\subsubsection{Analysis of the data augmentation}

\begin{figure}[H]
\centering
\includegraphics[width=0.6\textwidth]{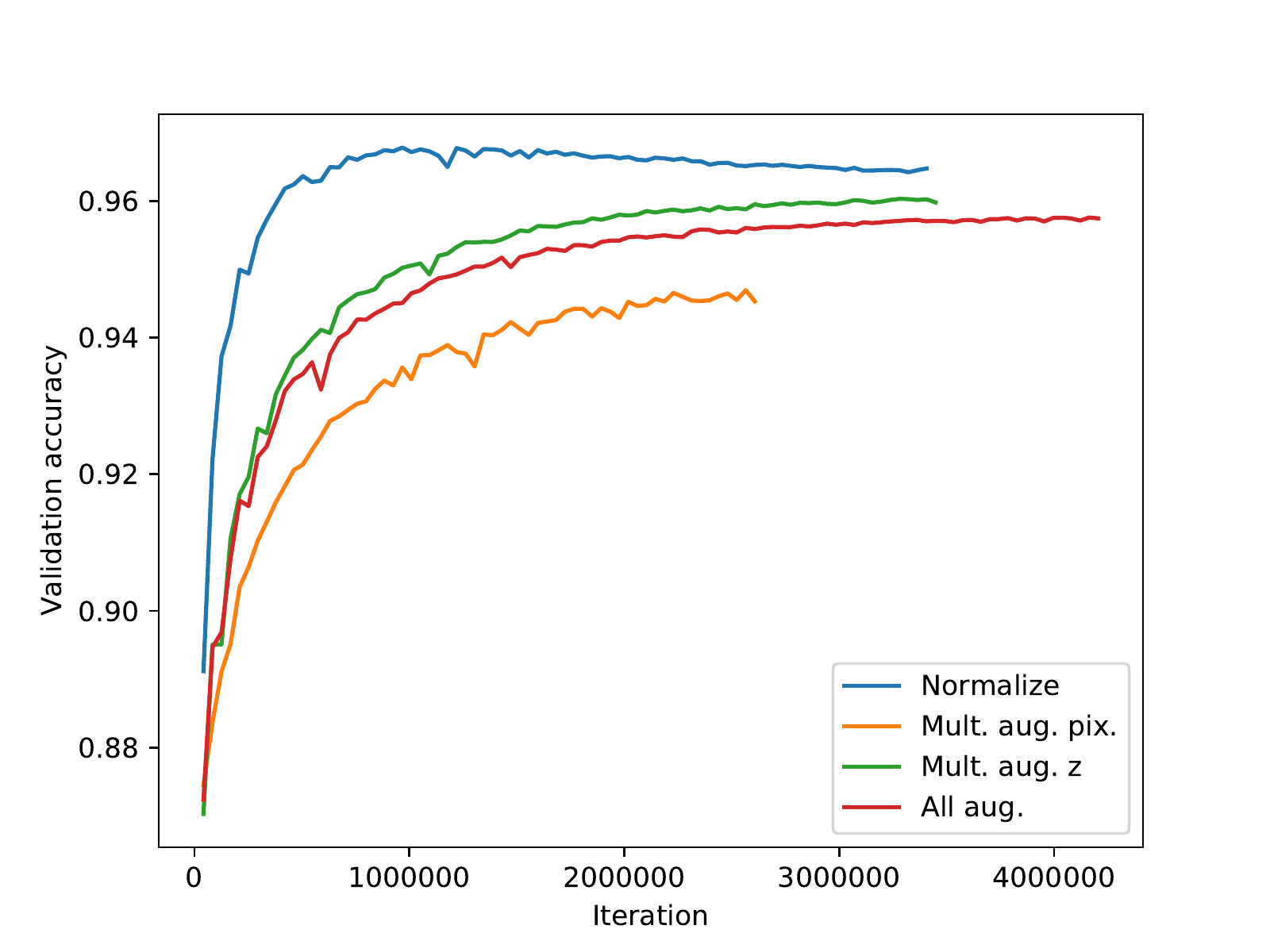}
\caption{Impact of different augmentations on the validation accuracy on the image split.}
\label{fig:image_split_plot}
\end{figure}

We analyze the impact of different augmentation procedures described
 in \autoref{sec:data_aug}.
\autoref{tab:augmentations} and \autoref{fig:image_split_plot} show the obtained validation accuracies
for different combinations of the augmentations when tested on the image split. 
 We observe that adding the additional augmentations reduces the validation accuracy on the image split, but also
 decreases the over-fit to the training data. 
The results also show that our proposed adjustment of the input depth variable $z$ along with the pixel values during multiplicative augmentation improves the
model performance. 

\begin{table}[H]
\centering
\begin{tabular}{c|c|c|c|c|c}  
Normalize & Symmetrize & Mult. Aug Pixels & Mult. Aug. Z (ours)& GP. Aug & Val. Acc        \\
\hline
 \checkmark        &            &                  &              &         & 96.78 \\ 
 \checkmark        & \checkmark          & \checkmark                &              &         & 94.64        \\
 \checkmark         & \checkmark          & \checkmark                & \checkmark            &         & 96.03           \\
 \checkmark         & \checkmark          & \checkmark                & \checkmark            & \checkmark       & 95.75
\end{tabular}
\caption{Impact of different data augmentations on the validation accuracy on the image split.}
\label{tab:augmentations}
\end{table}

\subsubsection{Impact on generalization}

To verify that the additional data augmentations improve the generalization, we compare the model with and without
the augmentations on all the three splits. \autoref{tab:splits_tab} 
shows that the augmentations in fact increase the validation accuracy on both the pose and the object split, while maintaining
the high image split performance. This indicates that the data augmentations result
in an increased generalization of the model.


\begin{table}[H]
\centering
\begin{tabular}{lccc}
              & \multicolumn{3}{c}{Train/validation split type} \\ \cline{2-4}
              & Image          & Pose          & Object         \\ \hline
Mahler et. al & 92.2           & 88.9          & 85.9           \\
Our GQ-CNN no aug.  & 96.7           & 88.2          & 86.7           \\
Our GQ-CNN all aug.  & 95.8           & 89.7          & 88.0
\end{tabular}
\caption{Validation accuracy on the different types of splits.}
\label{tab:splits_tab}
\end{table}


\subsubsection{Calibration}
Calibration indicates the relation between the predicted probability of success vs
ground-truth success proportion. The ground-truth success proportion is calculated as
the mean of success metric for predictions within a bucket with predictions within a given
success probability range. For instance, predictions in a bucket with success probability between
55\% and 65\% with perfect calibration would have a ground-truth mean success proportion of around 60\%.    

%

An accurate calibration of the prediction probabilities is crucial in the grasp planning systems that uses GQ-CNN.
The GQ-CNN prediction probabilities are used during an iterative sampling procedure to obtain the best grasp plan.
In consequence, a poor calibration of the predictions will result in a bad grasp plan.
Furthermore, it is well known that the modern deep neural networks often
obtain poorly calibrated probability estimates \cite{guo2017calibration}. Because of that, we investigate the calibration
of the obtained predictions and the impact of the data augmentations on the calibration.

\begin{figure}[H]
\centering
\includegraphics[width=0.6\textwidth]{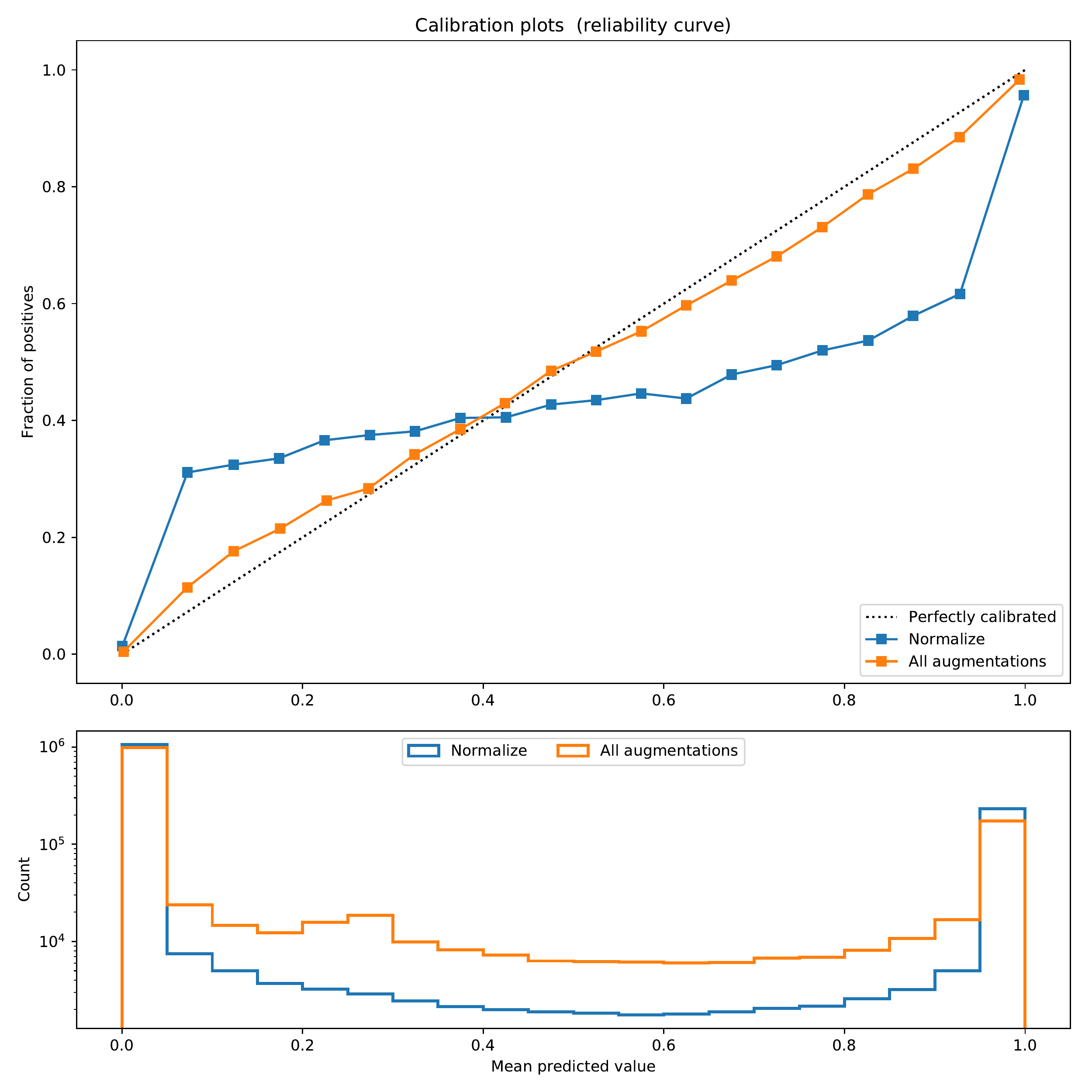}
\caption{Impact of the data augmentations on the calibration of the GQ-CNN predictions for the image split.
The upper subplot shows the calibration curves and the lower subplot shows counts of predictions
for each bucket. }
\label{fig:calibration}
\end{figure}

\autoref{fig:calibration} shows that the predictions obtained without using the data augmentations
show very poor calibration. On the other hand, when the data augmentations are used the
predictions are well calibrated. Therefore, without the data augmentations the final
 grasping performance might be worse due to the poorly calibrated grasp sampling procedure.

\section{Conclusions}

We proposed several improvements to the GQ-CNN grasp planning system.
We devised a new network architecture for GQ-CNN that outperforms the original one.
We enriched the data augmentation scheme which resulted in better
generalization on 
the more challenging train/test splits. Finally, we showed that the data augmentation
has a large impact on the calibration of the network predictions, which are crucial for
finding robust grasps.

In the future will perform experiments on a real robot to validate whether our
 improvements on the synthetic data are reflected in a better grasping performance.
Additionally, we are planning to further improve the GQ-CNN grasping system by
generating a larger and more diverse training dataset,
using higher resolution of the input images,
and increasing the sampling efficiency of the grasping policy. 

\section{Acknowledgements}
We thank Jeff Mahler for helpful discussions as well as
Ken Goldberg's Laboratory for Automation Science and Engineering at the UC Berkeley
for publishing their work on the GQ-CNN and hosting the benchmarks against our method.
We also thank NVIDIA for the Titan X graphic cards donated to the University of Warsaw.

\printbibliography

\end{document}